\documentclass{article}
\usepackage{ijcai25}
\usepackage{float}
\usepackage{times}
\usepackage{soul}
\usepackage{url}
\usepackage[hidelinks]{hyperref}
\usepackage[utf8]{inputenc}
\usepackage[small]{caption}
\usepackage{graphicx}
\usepackage{amsmath}
\usepackage{amsthm}
\usepackage{booktabs}
\usepackage{algorithm}
\usepackage[switch]{lineno}
\usepackage{amsfonts,amssymb}
\usepackage{color}
\usepackage{bm}
\usepackage{tcolorbox}

\usepackage{xspace}
\newcommand{\method}{\texttt{ARMR}\xspace }
\usepackage{multicol}
\usepackage{multirow}
\usepackage{algpseudocode}
\algnewcommand\algorithmicInput{\textbf{Input:}}
\algnewcommand\algorithmicOutput{\textbf{Output:}}
\algnewcommand\Input{\item[\algorithmicInput]}
\algnewcommand\Output{\item[\algorithmicOutput]}
\newcommand{\spara}[1]{\smallskip\noindent{\bf #1}}

\title{ARMR: Adaptively Responsive Network for Medication Recommendation}

\author{
Feiyue Wu$^{1}$
\and
Tianxing Wu$^{1,2}$\footnote{Corresponding author.} \and
Shenqi Jing$^{3}$\\
\affiliations
$^1$School of Computer Science and Engineering, Southeast University, Nanjing, China\\
 $^2$Key Laboratory of New Generation Artificial Intelligence Technology and Its Interdisciplinary Applications (Southeast University), Ministry of Education, China\\
$^3$The First Affiliated Hospital with Nanjing Medical University \\(JiangSu Province Hospital), Nanjing, China\\
\emails
\{wufeiyue, tianxingwu\}@seu.edu.cn, 
jingshenqi@jsph.org.cn
}

\begin{document}

\maketitle

\begin{abstract}
  Medication recommendation is a crucial task in healthcare, especially for patients with complex medical conditions. However, existing methods often struggle to effectively balance the reuse of historical medications with the introduction of new drugs in response to the changing patient conditions. In order to address this challenge, we propose an Adaptively Responsive network for Medication Recommendation (\method), a new method which incorporates 1) a piecewise temporal learning component that distinguishes between recent and distant patient history, enabling more nuanced temporal understanding, and 2) an adaptively responsive mechanism that dynamically adjusts attention to new and existing drugs based on the patient's current health state and medication history. Experiments on the MIMIC-III and MIMIC-IV datasets indicate that \method has better performance compared with the state-of-the-art baselines in different evaluation metrics, which contributes to more personalized and accurate medication recommendations. The source code is publicly avaiable at: \url{https://github.com/seucoin/armr2}.
\end{abstract}


\section{Introduction}
In the era of big data, abundant health information enables researchers and clinicians to develop sophisticated predictive methods for clinical decision support~\cite{doctorai,ehr_review,ali2023deep}. Medication recommendation aims to provide personalized and effective medication plans based on a patient's medical history, current diagnoses, and other relevant information. However, this task is particularly challenging for patients with complex conditions~\cite{cognet,gbert}, as their medical histories are often intricate and their responses to medications can be highly variable.

Deep learning methods have emerged as a powerful tool for modeling complex patterns in healthcare data~\cite{leap,dmnc,amanet}. Their ability to capture intricate patterns and dependencies~\cite{gamenet,premier,pmdc-rnn} makes them well-suited for medication recommendation tasks. Unlike traditional rule-based or instance-based methods, deep learning methods can learn from large datasets and adapt to the unique characteristics of individual patients. This adaptability is crucial for developing personalized medication plans that can evolve with the patient's changing health status. Despite the advantages of deep learning methods, existing methods in medication recommendation face several limitations:

\begin{figure}[tbp]
\centering
\includegraphics[width=\linewidth]{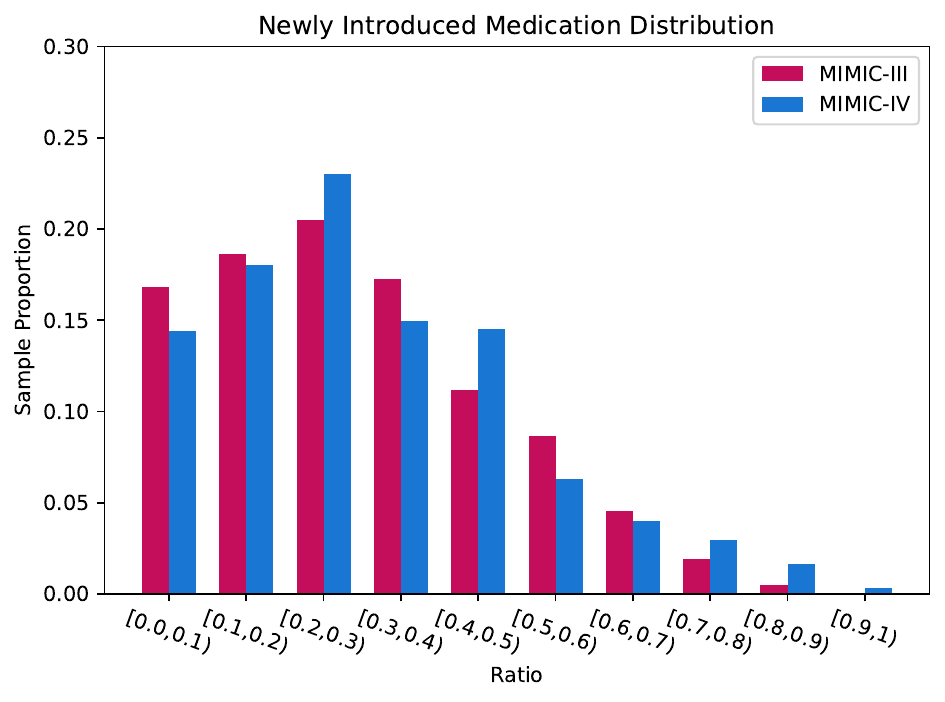}
\caption{
The proportion of newly introduced drugs in all drug prescriptions in the datasets. The x-axis represents the proportion range of new drugs, and the y-axis represents the proportion of prescriptions with the corresponding ratio in the dataset.
}
\label{fig:stats1}
\end{figure}

\spara{Neglecting the Importance of New Drug Patterns. } Figure~\ref{fig:stats1} shows the distribution of new and old drugs in prescribed medications. 
New drugs represent the drugs that has never been used before, while old drugs are the opposite.
Each time the prescribed drug set contains about 30\% new drugs on average. Many methods focus on modeling how to fully reuse old drugs~\cite{cognet,premier,micron,retain} or recommending full medication sets~\cite{leap,gamenet}, but neglect the importance of prescribing new drugs. For example, if a patient with hypertension is prescribed a new drug hydrochlorothiazide at some point, amlodipine is likely to be prescribed later. Learning the pattern of introduced newly prescribed drugs may help the method better understand the evolving health conditions as physicians adapt treatment plans to the patient's changing requirements.

\begin{figure}[tbp]
\centering
\includegraphics[width=\linewidth]{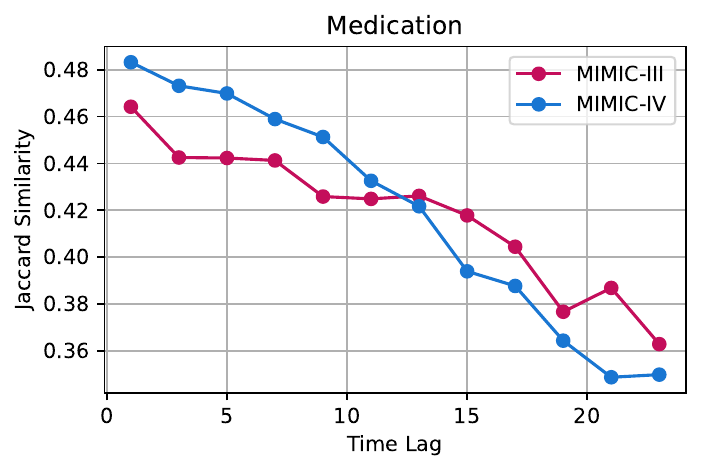}
\caption{
The similarity between two prescriptions for patients shows a decreasing trend with the increase of the interval between two admissions. The x-axis represents the interval between admission times for prescriptions, and the y-axis denotes the average similarity between all prescription pairs at the corresponding interval.
}
\label{fig:stats2}
\end{figure}

\spara{Insufficient Exploitation of History Records. } Figure~\ref{fig:stats2} shows a statistical analysis of the datasets. Generally, the larger the time lag between admission records, the greater the difference between medical code sets. Many existing methods focus primarily on recent patient data, neglecting the potential value of long-term historical information. However, distant records can still be crucial. For example, a patient's severe health event from several years ago may be crucial for understanding their current condition, yet traditional methods may overlook this due to their focus on recent records.

To address these limitations, we propose the Adaptively Responsive Network for Medication Recommendation (\method). This method adjusts attention to new and old drugs through an adaptively responsive mechanism, according to the patient's current health condition and drug usage history. Furthermore, we introduce a PTL component to effectively distinguish and process a patient's long-term and short-term health records, thereby improving the accuracy and personalization of medication recommendations. Our major contributions are:
\begin{itemize}
    \item  We propose a medication recommendation method that combines a dynamic adaptive mechanism and piecewise temporal learning. Compared to existing methods, our method is the first to explicitly distinguish the recommendation needs for new and history drugs, and dynamically adjust the recommendation strategy based on changes in the patient's current health state.
    \item  We develop a unique piecewise temporal learning component which can more accurately process and analyze information in long-term health records, providing more personalized and accurate medication recommendations.
    \item We conduct comprehensive experiments on two public medical datasets. \method outperforms the state-of-the-art baselines by 2.16\% improvement in Jaccard similarity and 2.55\% in PRAUC, respectively.
\end{itemize}

\section{Related Work}
In recent years, medication recommendation has garnered significant attention due to its potential to improve patient care. Existing methods can be broadly classified into three categories based on the information they utilize, which are rule-based methods, instance-based methods, and longitudinal methods.

\spara{Rule-based methods.} These methods typically rely on predefined clinical guidelines or attempt to derive rules from electrical health records~\cite{pilot,advisory}. For instance,~\cite{markov} employs Markov Decision Processes to learn mappings between patient characteristics and treatments, optimizing a series of if-then-else rules. Another study~\cite{rulebased} focuses on extracting rules from discharge summaries. However, these methods often require substantial input from healthcare professionals and may lack generalizability.

\spara{Instance-based methods.} These methods focus solely on current visit data to extract patient features.~\cite{leap} conceptualizes medication recommendations as a multi-instance multi-label task, proposing a sequence-to-sequence method with a content-attention mechanism.~\cite{smr} approaches the problem as a link prediction task, jointly embedding diseases, medicines, patients, and their relationships into a shared space using knowledge graphs. However, these methods often neglect valuable historical patient data.

\spara{Longitudinal methods.} This category represents popular methods that leverages sequential dependencies in patient treatment histories~\cite{retain,cognet,micron,carmen,molerec,leader_llm,gamenet,4sdrug,kgmt,gbert,refine}. Many of these methods employ RNN-based architectures to capture longitudinal patient information. For example,~\cite{retain} introduces a two-level neural attention network to effectively identify relevant information from patient visits.~\cite{gamenet} incorporates a memory bank to integrate global medication interaction knowledge. MICRON~\cite{micron} utilizes residual networks to predict health condition changes by analyzing differences between consecutive visits. Some studies~\cite{carmen,molerec} combine RNNs with graph neural networks to enhance medication representation. COGNet~\cite{cognet} employs Transformer architecture to encode medical code sets and introduces a copy module for explicit reuse of historical medications. More recently,~\cite{leader_llm} explores the potential of large language models for medication recommendation.

Our proposed method distinguishes itself from existing work by explicitly differentiating between historical and new medications and by incorporating fine-grained temporal learning to capture evolving patient conditions. This strategy aims to provide more precise and personalized medication recommendations.

\section{Preliminaries}

\subsection{Problem Formulation}

\spara{Definition 1 (Electrical Health Records)}
\textit{
Electrical Health Records (EHRs) collect the medical information of patients over time. An EHR is composed of a series of medical visits denoted as $\mathcal{V}^{(i)} = \{ \mathcal{V}^{(i)}_1, \mathcal{V}^{(i)}_2, \cdots, \mathcal{V}^{(i)}_{T^{(i)}} \}$, where $T^{(i)}$ represents the total number of visits for the $i$-th patient. For notational simplicity, we omit the patient index $i$ when the context is clear. }\textit{Each visit contains three essential sets: the diagnosis set $\mathcal{D} = \{ \mathbf{d}_1, \mathbf{d}_2, \cdots, \mathbf{d}_{|\mathcal{D}|} \}$, the procedure set $\mathcal{P} = \{ \mathbf{p}_1, \mathbf{p}_2, \cdots, \mathbf{p}_{|\mathcal{P}|} \}$, and the medication set $\mathcal{M} = \{ \mathbf{m}_1, \mathbf{m}_2, \cdots, \mathbf{m}_{|\mathcal{M}|} \}$. Here, $|\cdot|$ denotes the cardinality of each set. Each visit $\mathcal{V}_t$ is then characterized by a triplet $\{ \mathcal{D}_t, \mathcal{P}_t, \mathcal{M}_t \}$, where $\mathcal{D}_t \subseteq \mathcal{D}$, $\mathcal{P}_t \subseteq \mathcal{P}$, and $\mathcal{M}_t \subseteq \mathcal{M}$ represent the diagnoses, procedures, and medications associated with that visit, respectively.
}

\spara{Definition 2 (Medication Recommendation)}
\textit{
Given a patient's current diagnoses $\mathcal{D}_t$, procedures $\mathcal{P}_t$, and historical visit information $\{ \mathcal{V}_1, \mathcal{V}_2, \cdots, \mathcal{V}_{t-1} \}$, medication recommendation aims to recommend an appropriate medication combination $\mathcal{M}_t$ for the patient's current visit.
}

\begin{table}[ht]
\centering
\begin{tabular}{l|l}

\toprule
\textbf{Notation}   &   \textbf{Description} \\
\midrule
\midrule
$\mathcal{V}$   &   The record for a single patient \\
$\mathcal{D}$   &   All diagnosis set \\
$\mathcal{P}$   &   All procedure set \\
$\mathcal{M}$   &   All medication set \\
T   & The length of patient's visit records \\
$\mathbf{d}$   &   The diagnosis set of a single visit \\
$\mathbf{p}$   &   The procedure set of a single visit \\
$\mathbf{m}$   &   The medication set of a single visit \\
\bottomrule

\end{tabular}
\caption{The notations used in \method.}
\label{tab:notation}
\end{table}

\section{The Proposed Method}

\begin{figure*}[t]
\centering
\includegraphics[width=1.0\textwidth]{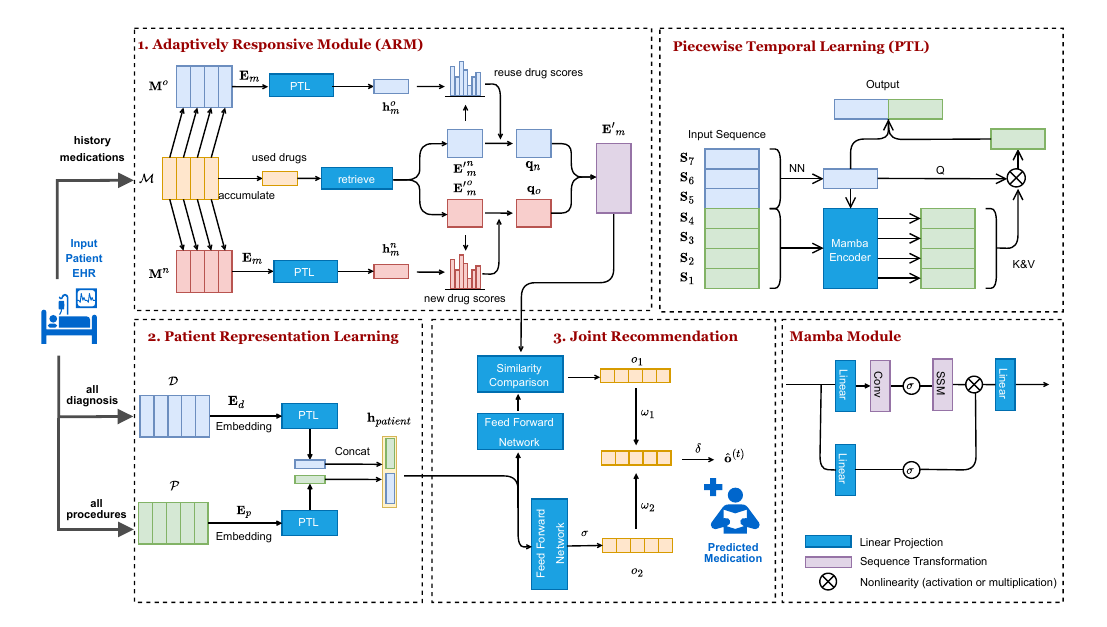}
\caption
{
The overview of \method's architecture. Given the input longitudinal patient EHR which comprises $\mathbf{M}$, $\mathbf{D}$ and $\mathbf{P}$, we predict the last medication set $\hat{\mathbf{o}}^{(t)}$. \textbf{Step 1:} learning personalized medication representation $\mathbf{E'}_m$, \textbf{Step 2:} learning the patient's comprehensive representation $\mathbf{h}_{patient}$, \textbf{Step 3: } obtaining the final predicted medication set $\hat{\mathbf{o}}^{(t)}$. The Mamba module is utilized in PTL, while the PTL component is used as a sequence temporal extractor.
}
\label{fig:framework}
\end{figure*}

\subsection {Overall Arcitecture}
As shown in Figure~\ref{fig:framework}, our \method method comprises one pluggable component and three modules: (1) a hybrid \textbf{ piecewise temporal learning} (PTL) component that encodes temporal medical codes set sequences by considering two stages; (2) an \textbf{adaptively responsive module} (ARM) that learns dynamic medication representation from both historical medication usage and new medication in response to changing conditions; (3) a \textbf{ patient representation learning module  } that learns patients' comprehensive representations from their EHR data; (4) the patient's and medication's representation vectors are combined in the \textbf{ joint recommendation module}, where the final drug output is obtained from element-wise thresholding.

\subsection{Piecewise Temporal Learning}

The piecewise temporal learning component addresses the varying impact of long-term and short-term health records on drug recommendations. It divides health records into ``recent'' and ``distant'' segments. Given an input sequence $\mathbf{S} \in \mathbb{R}^{T\times dim}$ of length $T$ $(T \geq N)$, we split it into a recent part: $\mathbf{S}_{near} = [\mathbf{S}^{(1)}, \mathbf{S}^{(2)}, \dots, \mathbf{S}^{(N)}] \in \mathbb{R}^{N\times dim}$ and a distant part: $\mathbf{S}_{far} = [\mathbf{S}^{(N+1)}, \mathbf{S}^{(N+2)}, \dots, \mathbf{S}^{(T)}] \in \mathbb{R}^{(T-N)\times dim}$, and $N$ is a hyperparameter determining the split point.
%
For $\mathbf{S}_{near}$, we apply a feedforward neural network to capture recent health changes $\mathbf{h}_{near}$ as follows:
\begin{equation}
\mathbf{h}_{near} = \mathrm{LayerNorm}(\mathbf{S}_{near} + \mathrm{FF_{PTL}}(\mathbf{S}_{near}))
\end{equation}
where $\mathrm{LayerNorm}$~\cite{layernorm} is layer normalization function, and $\mathrm{FF_{PTL}}$ is a two-layer fully connected feedforward network. 
For distant health changes, we need an efficient sequence modeling architecture that can effectively capture long-range dependencies while maintaining computational efficiency. Mamba~\cite{mamba}, a state-of-the-art selective state space model, has shown superior performance in modeling long sequences through its input-dependent selection mechanism. While transformer-based (with quadratic-complexity) architectures are possible alternatives, Mamba offers comparable sequence modeling capabilities with significantly better computational efficiency (near linear). This makes it particularly suitable for processing longitudinal patient records with long-term patterns. Thus, we utilize the $\mathrm{Mamba}$ block to obtain distant health changes representation $\mathbf{h}_{far}$ as follows:
\begin{equation}
\label{exp:ptl.far}
\mathbf{h}_{far} = \mathrm{Mamba}(\mathbf{S}_{far}^{(1)}, \dots, \mathbf{S}_{far}^{(T-N)})
\end{equation}
%

We then combine the recent health change representation $\mathbf{h}_{near}$ and the distant health change representation $\mathbf{h}_{far}$ to get an enhanced distant health change representation $\mathbf{h'}_{far}$ using an attention mechanism as follows:
\begin{equation}
\mathbf{h'}_{far} = \mathrm{LayerNorm}(\mathbf{h}_{near} + \mathrm{Softmax}(\mathbf{h}_{near}(\mathbf{h}_{far})^T)\mathbf{h}_{far})
\end{equation}
where $\mathrm{Softmax}$ is a commonly used mathematical function to transform a vector of values into a probability distribution.
This operation aims to make the component focus on relevant historical information based on recent context.
Finally, we concatenate the processed representations to get an overall temporal representation $\mathbf{h}_S$ as follows:

\begin{equation}
\mathbf{h}_{S} = \mathrm{CONCAT}(\mathbf{h}_{near}, \mathbf{h'}_{far}) \in \mathbb{R}^{2N \times dim}
\end{equation}
where $\mathrm{CONCAT}$ concatenation between vectors. We denote this entire process as $\mathrm{PTL}(\mathbf{S}) = \mathbf{h}_S$.
%
The PTL component integrates multiple techniques to analyze temporal data by distinguishing between recent and historical information. In this component, the feedforward network is utilized to capture immediate health status, while the Mamba block is employed to learn distant health status.

\subsection{Adaptively Responsive Module}

The adaptively responsive module generates dynamic medication embeddings based on the patterns of newly introduced drugs and historically reused drugs. By distinguishing both types of drugs, we aim to personalize recommendations to make the module adapt to both continuing treatments and newly introduced medications.
Given a sequence of drug sets $\{\mathbf{m}_1, \mathbf{m}_2, \ldots, \mathbf{m}_{T-1}\}$ where $\mathbf{m}_i \subseteq \mathcal{M}$, we define two set sequences, i.e., the existing drug set sequence: $\{\mathbf{m}_1^o, \mathbf{m}_2^o, \ldots, \mathbf{m}_{T-1}^o\}$ and the new drug set sequence: $\{\mathbf{m}_1^n, \mathbf{m}_2^n, \ldots, \mathbf{m}_{T-1}^n\}$ as follows:

\begin{equation}
\mathbf{m}_i^o =
\begin{cases}
\emptyset & \text{if } i = 1 \\
\mathbf{m}_i \cap \left(\bigcup_{k=1}^{i-1} \mathbf{m}_k\right) & \text{if } i > 1
\end{cases}
\end{equation}

\begin{equation}
\mathbf{m}_i^n =
\begin{cases}
\mathbf{m}_i & \text{if } i = 1 \\
\mathbf{m}_i \setminus \mathbf{m}_i^o & \text{if } i > 1
\end{cases}
\end{equation}
where $\mathbf{m}_i^o$ represents the medications that were also prescribed in previous visits, while $\mathbf{m}_i^n$ represents newly introduced medications at each visit. We use $\cap$ to represent set intersection, and $\setminus$ to represent set difference. The separation of the existing drug set and the new drug set for each prescription aims to make the module differentiate between ongoing treatments and new interventions.
%
We embed these sequences and apply two different PTL components $\mathrm{PTL_m^o}$ and $\mathrm{PTL_m^n}$ corresponding to $\mathbf{m}_i^o$ and $\mathbf{m}_i^n$ respectively, and then obtain time-sensitive representations $\mathbf{h}_m^o$ and $\mathbf{h}_m^n$ as follows:

\begin{equation}
\mathbf{h}_m^o = \mathrm{PTL_m^o}(\{\mathbf{m}_i^o\mathbf{E}_m|1\le i < T\})
\end{equation}

\begin{equation}
\mathbf{h}_m^n = \mathrm{PTL_m^n}(\{\mathbf{m}_i^n\mathbf{E}_m|1\le i < T\})
\end{equation}
where $\mathbf{E}_m \in \mathbb{R}^{|\mathcal{D}|\times dim}$ is the base medication embedding which is randomly initialized. The use of PTL aims to make the module capture temporal patterns in both existing and new medication sequences.
We then define two medication set masks: used drugs mask $\mathbf{h}^\mathcal{O} \in \mathbb{R}^{|\mathcal{M}|}$ and unused drugs mask $\mathbf{h}^\mathcal{N} \in \mathbb{R}^{|\mathcal{M}|}$, to explicitly select the used drugs and unused drugs, respectively. These are derived from the historical medication sequence $\mathbf{m}_h = \bigcup_{i=1}^{T-1} \mathbf{m}_i$. $\mathbf{h}^\mathcal{O}_j$ and $\mathbf{h}^\mathcal{N}_j$ are defined as the $j$-th values of $\mathbf{h}^\mathcal{O}$ and $\mathbf{h}^\mathcal{N}$ respectively:

\begin{equation}
\mathbf{h}^\mathcal{O}_j = \sum_{i=1}^{T-1} ( \alpha_i \cdot e^{-\beta_i \cdot (T-i)})*\mathbf{1}\{{\mathcal{M}_j \in \mathbf{m}_h}\}, 1 \le j \le |\mathcal{M}|
\end{equation}

\begin{equation}
\mathbf{h}^\mathcal{N}_j = 1 - \mathbf{1}\{{\mathcal{M}_j \in \mathbf{m}_h}\}, 1 \le j \le |\mathcal{M}|
\end{equation}
where $\alpha_i$ and $\beta_i$ control time decay. $\mathbf{h}^\mathcal{O}_j$ leverages the occurrence time, frequency, and time decay of medications to build the representation of medication history. $\mathbf{1}\{\cdot\}$ is an indicator function which will return 1 if the given expression is true and otherwise 0. $\mathbf{h}^\mathcal{N}_j$ indicates unused medications, aiming to make the module consider potential new treatments.
Finally, we compute the responsive medication representation $\mathbf{E'}_m \in \mathbb{R}^{|\mathcal{D}|\times dim}$ as follows:

\begin{equation}
\begin{split}
\label{exp:medrep}
& \mathbf{E'}_m =  \mathbf{E}_m \odot (\mathbf{h}^\mathcal{O}\odot \mathbf{q}_o + \mathbf{h}^\mathcal{N} \odot \mathbf{q}_n) \\
& {\mathbf{q}}_o = \mathrm{Softmax}(\frac{\mathbf{E}_m(\mathbf{h}_m^o)^T\mathbf{W}_o}{\sqrt{dim}}) \\
& {\mathbf{q}}_n = \mathrm{Softmax}(\frac{\mathbf{E}_m(\mathbf{h}_m^n)^T\mathbf{W}_n}{\sqrt{dim}})
\end{split}
\end{equation}
where $\mathbf{W}_o, \mathbf{W}_n \in \mathbb{R}^{2N \times dim}$ are learned weights, and ${\mathbf{q}}_o$ and ${\mathbf{q}}_n$ combine the base medication embedding $\mathbf{E}_m$ with the time-sentisive representations $\mathbf{h}_m^o$ and $\mathbf{h}_m^n$ respectively, so that we can get a dynamic representation of each medication.


\subsection{Patient Representation Learning}
The patient representation learning module aims to create comprehensive representations of patients' health status based on their historical diagnosis and procedure codes. This module processes the visit records $\mathcal{V} = \{ \mathcal{V}_1, \mathcal{V}_2, \cdots, \mathcal{V}_T\}$ to encode both diagnosis and procedure information for each visit.
At first, we utilize an embedding table $\mathbf{E}_d \in \mathbb{R}^{|\mathcal{D}|\times dim}$, where each row represents a diagnosis code embedding. For a multi-hot diagnosis vector $\mathbf{d}^{(t)}\in \{0,1\}^{|\mathcal{D}|}$ at the $t$-th visit, we compute the diagnosis representation $\mathbf{d}^{(t)}_e$ as follows:

\begin{equation}
\mathbf{d}^{(t)}_e = \mathbf{d}^{(t)} \mathbf{E}_d
\end{equation}
This operation sums the embeddings of all active diagnosis codes for the visit.
Similarly, we use a procedure embedding table $\mathbf{E}_p \in \mathbb{R}^{|\mathcal{P}|\times dim}$ to encode the procedure vector $\mathbf{p}^{(t)}$ as follows:
\begin{equation}
\mathbf{p}^{(t)}_e = \mathbf{p}^{(t)} \mathbf{E}_p
\end{equation}
The resulting embedding vectors $\mathbf{d}^{(t)}_e$, $\mathbf{p}^{(t)}_e \in \mathbb{R}^{dim}$ encode the patient's health condition for each visit.
To capture longitudinal information, we apply two PTL components $\mathrm{PTL_d}$ and $\mathrm{PTL_p}$ to the sequences of diagnosis and procedure embeddings respectively as follows:

\begin{equation}
\begin{split}
& \mathbf{d}_h = \mathrm{PTL_d}(\{\mathbf{d}_e^{(t)}|1\le t \le T\}), \\
& \mathbf{p}_h = \mathrm{PTL_p}(\{\mathbf{p}_e^{(t)}|1\le t \le T\})
\end{split}
\end{equation}
where $\mathbf{d}_h, \mathbf{p}_h \in \mathbb{R}^{2N \times dim}$. This step allows the method to learn temporal patterns in both diagnosis and procedure histories.
Finally, We combine the diagnosis and procedure information to create the final patient representation as follows:
\begin{equation}
\mathbf{h}_{patient} = \mathrm{CONCAT}(\mathbf{d}_h, \mathbf{p}_h)
\end{equation}
where $\mathbf{h}_{patient} \in \mathbb{R}^{4N \times dim}$.
By integrating both the content (diagnoses and procedures) and the temporal aspects of a patient's medical history, this module creates a representation that can be used in the subsequent drug recommendation.

\subsection {Joint Recommendation}
%
At first, we employ a direct mapping from the patient representation to drug recommendations. This is achieved through a specialized neural network $\mathrm{FF}_{o_1}$, which transforms the patient encoding $\mathbf{h}_{patient}$ into a preliminary recommendation vector $\mathbf{o}_1$ as follows:
\begin{equation}
\mathbf{o}_1=\mathrm{FF_{o_1}}(\mathbf{h}_{patient}), \quad \mathbf{o}_1 \in \mathbb{R}^{|\mathcal{M}|}
\end{equation}

Concurrently, we utilize a similarity-based method to capture nuanced relationships between patient characteristics and drug properties to obtain $\mathbf{o}_2$. This involves projecting the patient representation into a drug-compatible space using another neural network $\mathrm{FF}_{o_2}$, followed by a similarity computation with drug embeddings $\mathbf{E'}_m$ as follows:
\begin{equation}
\mathbf{o}_2=\mathrm{sim}(\mathrm{FF}_{\mathbf{o}_2}(\mathbf{h}_{patient}),\mathbf{E'}_m), \quad \mathbf{o}_2 \in \mathbb{R}^{|\mathcal{M}|}
\end{equation}
%
%
where $\mathrm{sim}(\cdot,\cdot)$ represents the cosine similarity function as follows:
\begin{equation}
\mathrm{sim}(\mathbf{x}, \mathbf{y}) = \frac{\mathbf{x} \cdot \mathbf{y}}{|\mathbf{x}| |\mathbf{y}|}
\end{equation}

The final recommendation $\hat{\mathbf{o}}$ is derived by combining these two outputs $\mathbf{o}_1$ and $\mathbf{o}_2$ through a weighted sum, followed by a sigmoid activation to ensure output values are constrained between 0 and 1 as follows:
\begin{equation}
\label{eq:final}
\hat{\mathbf{o}} = \sigma(\mathbf{w}_1 \mathbf{o}_1 + \mathbf{w}_2 \mathbf{o}_2)
\end{equation}
where $\mathbf{w}_1$ and $\mathbf{w}_2$ are learnable parameters that determine the relative importance of each recommendation, and $\sigma(\cdot)$ denotes the sigmoid function.


\subsection {Model Training and Inference}
This study frames the drug recommendation task as a multi-label binary classification problem. Let $\mathbf{m}^{(t)}$ represent the ground truth drug recommendation vector, and $\hat{\mathbf{o}}^{(t)}$ is a vector denoting the method's output probalilities for each drug (as opposed to the final binary predictions $\hat{\mathbf{m}}^{(t)}$ obtained after thresholding).
Our training strategy employs two complementary loss functions. First, we utilize the binary cross-entropy loss $\mathcal{L}_{bce}$, treating each drug prediction as an independent sub-problem as follows:

\begin{equation}
\mathcal{L}_{bce} =
-\sum_{i=1}^{|\mathcal{M}|} \mathbf{m}^{(t)}_i log (\hat{\mathbf{o}}^{(t)}_i) + (1-\mathbf{m}^{(t)}_i)log (1-\hat{\mathbf{o}}^{(t)}_i)
\end{equation}
where $\mathbf{m}^{(t)}_i$ and $\hat{\mathbf{o}}^{(t)}_i$ are the $i$-{th} entries. 
To enhance the robustness of our predictions, we incorporate a multi-label hinge loss $\mathcal{L}_{multi}$, which ensures that the scores for true positive labels maintain a minimum margin above those of true negative labels as follows:
\begin{equation}
\mathcal{L}_{multi} = \sum_{i,j:~\mathbf{m}^{(t)}_i=1,\mathbf{m}^{(t)}_j= 0}
\frac{\mbox{max}(0,1-(\hat{\mathbf{o}}^{(t)}_i-\hat{\mathbf{o}}^{(t)}_j))}{|\mathcal{M}|} 
\end{equation}

The overall loss is computed as a weighted combination~\cite{loss_sum} of the above two losses as follows:
\begin{equation}  \label{eq:alpha}
\mathcal{L} = \alpha \mathcal{L}_{bce} +(1-\alpha)\mathcal{L}_{multi} 
\end{equation}
where $\alpha \in [0,1]$ controls the relative importance of each loss term, and we emprically set $\alpha = 0.7$.
During training, we aggregate the loss across all visits for each patient before performing backpropagation. This patient-level optimization allows the method to capture long-term dependencies in medication histories.
For inference, we follow a similar process in the training phase. The final drug recommendations are determined by applying a threshold $\delta$ to the output probabilities $\hat{\mathbf{o}}^{(t)}$. Specifically, we recommend drugs corresponding to the entries where $\hat{\mathbf{o}}^{(t)} > \delta$, and $\delta$ is emprically set as 0.5.

\section{Experiments}
This section presents our experimental results, demonstrating the efficacy of our proposed \method method.
\begin{table}[t]
	\centering
		\begin{tabular}{l|l|l}
			\toprule
			\textbf{Item} & \textbf{MIMIC-III} & \textbf{MIMIC-IV}\\
			\midrule
			\midrule
			\# of patients & 6,350 & 9,036 \\
			\# of visits & 15,032 & 20,616\\
			avg. \# of visit per patient & 2.37 & 2.28\\
			max. \# of visit & 29 & 28\\
                \# of diagnosis codes & 1,958 & 1,892\\
                \# of procedure codes & 1,430 & 4,939\\
                \# of medication codes & 131 & 131\\
			\bottomrule
		\end{tabular}
        \caption{The statistics of the processed dataset.}
        \label{tab:data_stats}
\end{table}

\begin{table*}[ht]
\centering
\begin{tabular}{|c|ccc|ccc|}
\hline
\multirow{2}{*}{Method} & \multicolumn{3}{c|}{MIMIC-III} & \multicolumn{3}{c|}{MIMIC-IV} \\ \cline{2-7} 
                        & Jaccard (\%)   & PRAUC (\%)    & F1 (\%) & Jaccard (\%)   & PRAUC (\%)    & F1 (\%) \\ 
\hline
\hline
LR                      & 49.40$\pm$0.14 & 75.88$\pm$0.19 & 65.08$\pm$0.11 & 47.32$\pm$0.18 & 73.80$\pm$0.16 & 63.03$\pm$0.17 \\ 
RETAIN                  & 48.55$\pm$0.15 & 75.68$\pm$0.12 & 64.67$\pm$0.10 & 44.03$\pm$0.10 & 71.40$\pm$0.11 & 60.13$\pm$0.08 \\ 
LEAP                    & 45.44$\pm$0.18 & 65.71$\pm$0.12 & 60.57$\pm$0.12 & 43.50$\pm$0.10 & 71.61$\pm$0.40 & 60.13$\pm$0.23 \\ 
GAMENet                 & 51.59$\pm$0.09 & 76.11$\pm$0.07 & 66.74$\pm$0.10 & 48.84$\pm$0.08 & 73.95$\pm$0.05 & 64.41$\pm$0.17 \\ 
LEADER                  & 51.75$\pm$0.22 & 77.95$\pm$0.25 & 67.37$\pm$0.19 & 47.84$\pm$0.08 & 74.95$\pm$0.05 & 63.32$\pm$0.17 \\ 
Carmen                  & 52.67$\pm$0.21 & 76.52$\pm$0.36 & 68.12$\pm$0.19 & 50.06$\pm$0.12 & 74.62$\pm$0.30 & 65.69$\pm$0.07 \\ 
MoleRec                 & 53.05$\pm$0.33 & 77.36$\pm$0.27 & 68.43$\pm$0.29 & 49.92$\pm$0.21 & 74.73$\pm$0.31 & 65.45$\pm$0.19 \\
COGNet                  & 53.36$\pm$0.11 & 77.39$\pm$0.09 & 68.69$\pm$0.10 & 50.23$\pm$0.12 & 75.21$\pm$0.21 & 65.82$\pm$0.14 \\
\hline  
\textbf{\method}            & \textbf{54.51$\pm$0.13} & \textbf{79.63$\pm$0.28} & \textbf{69.59$\pm$0.29} & \textbf{51.74$\pm$0.14} & \textbf{76.95$\pm$0.21} & \textbf{67.12$\pm$0.18} \\
\method \textit{w/o PTL}    & 52.45$\pm$0.12 & 77.49$\pm$0.09 & 66.88$\pm$0.24 & 49.82$\pm$0.21 & 74.51$\pm$0.34 & 65.47$\pm$0.23 \\
\method \textit{w/o PTL(L)} & 54.01$\pm$0.08 & 78.99$\pm$0.18 & 69.32$\pm$0.16 & 51.32$\pm$0.17 & 76.46$\pm$0.21 & 66.58$\pm$0.15 \\
\method \textit{w/o PTL(N)} & 53.55$\pm$0.17 & 78.49$\pm$0.15 & 68.72$\pm$0.14 & 50.51$\pm$0.17 & 75.35$\pm$0.33 & 65.98$\pm$0.09 \\
\method \textit{w/o ARM}    & 53.21$\pm$0.14 & 78.28$\pm$0.34 & 68.69$\pm$0.30 & 50.22$\pm$0.10 & 75.20$\pm$0.11 & 65.66$\pm$0.14 \\
\hline
\end{tabular}
\caption{The comparison results on MIMIC-III and MIMIC-IV datasets.}
\label{tab:perf}
\end{table*}

\subsection{Experimental Setting}
\spara{Datasets.} 
We conducted experiments using the MIMIC-III~\cite{mimic3} and MIMIC-IV~\cite{mimic4} datasets. 
Following data preprocessing procedures outlined in~\cite{carmen}, we identified 131 medications for recommendation, using ATC Third Level codes as target labels.
Each ATC Third Level code encompasses one or more medications, with each medication corresponding to a single ATC Third Level code. 
%
%
Table~\ref{tab:data_stats} provides key statistics of the processed datasets.

\spara{Evaluation Metrics.} According to the previous studies in medication recommendation~\cite{safedrug,cognet,carmen}, we employed three primary metrics: 
1) Jaccard Similarity Score (Jaccard) quantifies the degree of overlap between two sets by computing the ratio of their intersection to their union; 
2) F1-score (F1) is the harmonic mean of precision and recall; 
3) PRAUC refers to precision recall area under curve. 
These metrics were calculated as the mean across all patients in the dataset.

\spara{Baselines.}
We compared our method with different baseline methods:
{\sloppy
1) Traditional machine learning methods like Logistic Regression (\textbf{LR});
2) Deep learning baselines including \textbf{RETAIN}~\cite{retain}, \textbf{LEAP}~\cite{leap}, \textbf{GAMENet}~\cite{gamenet}, \textbf{LEADER}~\cite{leader_llm}, \textbf{Carmen}~\cite{carmen}, \textbf{MoleRec}~\cite{molerec} and \textbf{COGNet}~\cite{cognet}.} Additionally, we evaluated several variants of our \method method which are detailed in the section of ablation study.

\subsection {Performance Comparison}

Table~\ref{tab:perf} presents a comprehensive comparison of all baseline methods and \method variants across both MIMIC-III and MIMIC-IV datasets.

\spara{Overall Performance.} Our \method method consistently outperforms all baseline methods in the metrics of Jaccard, F1 and PRAUC.
The simpler methods (LR, ECC, LEAP) show limited effectiveness, likely due to their focus on current visit data without considering longitudial patient history.
More sophisticated methods like RETAIN, GAMENet, MoleRec, Carmen and COGNet demonstrate improved performance through various methods of incorporating historical patient data. However, these methods still fall short in certain aspects.
RETAIN relies solely on bidirectional RNN for historical encoding.
While GAMENet introduces graph-based information but lacks in temporal dynamics 
and MoleRec incorporates drug molecule structures, enhancing performance but missing broader contextual factors. 
COGNet employs positional encoding self-attention, recognizing the potential for medication reuse, but overlooks the dynamic nature of health conditions.
In contrast, \method 's superior performance can be attributed to its comprehensive strategy in integrating historical data with current health dynamics.

\subsection{Ablation Study}
To evaluate the contribution of each component in our proposed \method method, we conducted a series of ablation experiments. We designed the following variant methods:
\begin{itemize}
    \item \method \textit{w/o PTL}: We used standard RNN instead of PTL.
    \item \method \textit{w/o PTL(L)}: We maintained the PTL module but removes the distant history processing part, which means simply changing Eq.(\ref{exp:ptl.far}) to an empty sequence.
    \item \method \textit{w/o PTL(N)}:  We maintained the PTL module but remove the recent history processing part by changing the input of Eq.(\ref{exp:ptl.far}) to the whole sequence and removing recent hisotry processing part.
    \item \method \textit{w/o ARM}:  We removed the ARM module by using PTL on the ${\mathbf{m}_i}$ sequence directly and get the final medication representations similarly to Eq.(\ref{exp:medrep})
\end{itemize}
%
The superior performance of \method over \method \textit{w/o PTL} and \method \textit{w/o ARM} indicates that the PTL component and ARM module bring significant improvement to the basic method without PTL and ARM. The PTL module creates a more accurate and context-aware representation of a patient's health history, potentially leading to more informed drug recommendations. The ARM module dynamically adjusts drug recommendations based on the patient's medication history and current health condition. By distinguishing between existing and new medications, and considering their temporal patterns, the module can provide a more personalized and context-aware recommendation strategy. 
The difference between \method \textit{w/o PTL} and \method \textit{w/o PTL(L)} is that they are using different base sequence learning architectures, i.e, standard RNN and Mamba. The superior performance of the latter highlights the advantages of using Mamba for medical sequence modeling, particularly in capturing long-range dependencies while maintaining computational efficiency.
Both \method \textit{w/o PTL(N)} and \method \textit{w/o PTL(L)} show suboptimal performance, emphasizing the importance of our temporal segmentation strategy in the PTL component. This suggests that balancing the influence of recent and distant history is critical for accurate medication recommendation.
In general, the complete \method outperforms all variants removing some part, confirming that each key design contributes synergistically to the overall performance.

\subsection {Case Study}
To demonstrate the efficacy of \method in balancing the continuation of previous treatments and the introduction of new medications, we conducted an in-depth analysis of a representative patient from the MIMIC-IV dataset. This individual's medical history comprises five distinct hospital visits, which are characterized by the evolving diagnoses and corresponding adjustments to their medication prescriptions.
\begin{figure}[ht]
\includegraphics[width=\linewidth]{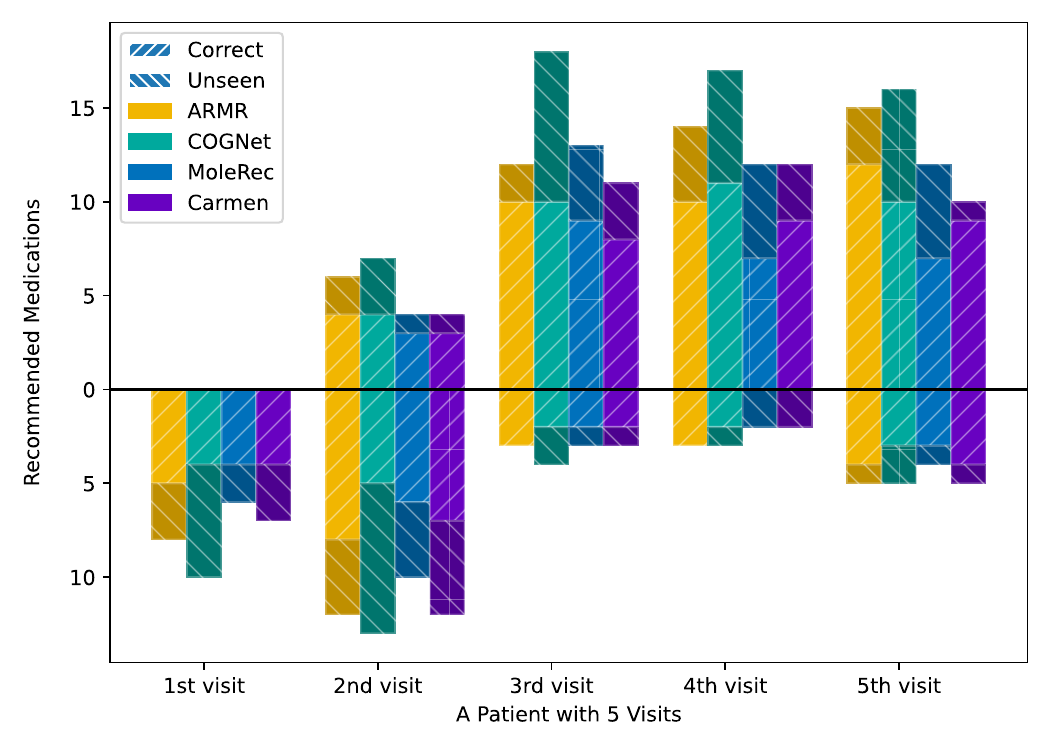}
\caption{An example of case study. The figure shows four methods' predicted medications for a patient with 5 visits. The y-axis shows the number of recommend drugs for each method per visit, which are divided into historical (upper) and new (lower) drugs split by the y=0 line. The bright color means true positives, and the dark color means false positives. }
\label{fig:case_study}
\end{figure}

Figure~\ref{fig:case_study} presents the medication recommendation results for the same patient generated by our method and three state-of-the-art baselines (i.e., COGNet, MoleRec and Carmen) which are good in general.
A comprehensive evaluation of the results reveals that \method consistently outperforms its counterparts in terms of overall accuracy of medication recommendations. Notably, \method recommends more correct historical drugs and relatively less unseen drugs, exhibiting comparable proficiency to COGNet in identifying appropriate medications for continued use, but COGNet tends to recommend more unseen drugs. Besides, \method predicts the most correct new drugs in all visits, which truly distinguishes itself for its superior ability to suggest new, previously unprescribed medications that align with the patient's evolving medical needs. These findings highlight the efficacy of our proposed ARM, which effectively balances continuity in treatment while addresses emergent patient needs.

\section{Conclusion}


In this paper, we propose a new medication recommendation method \method which leverages a piecewise temporal learning component to extract changing patient conditions, and utilizes an adaptively responsive module to determine how to use existing and new medications.
Experiment results on the publicly available MIMIC-III and MIMIC-IV dataset demonstrate that \method outperforms existing baselines.
In the future, we plan to 1) design quantified experiments to evaluate the specific ability of all methods on introducing correct new drugs, and 2) enhance the explainability of recommendations to further increase clinicians' trust to apply \method in real-world scenarios.

\section*{Acknowledgements}
This work is supported by the NSFC (Grant No. 62376058), the National Key Research and Development Program (Grant No. 2022YEB2703300, 2022YFB2703301), the Key R\&D Program of Jiangsu (Grant No. BE2022798), the Southeast University Interdisciplinary Research Program for Young Scholars, and the Big Data Computing Center of Southeast University.

\bibliographystyle{named}
\bibliography{ijcai25}

\begin{thebibliography}{}

\bibitem[\protect\citeauthoryear{Abelson \bgroup \em et al.\egroup
  }{1985}]{abelson-et-al:scheme}
Harold Abelson, Gerald~Jay Sussman, and Julie Sussman.
\newblock {\em Structure and Interpretation of Computer Programs}.
\newblock MIT Press, Cambridge, Massachusetts, 1985.

\bibitem[\protect\citeauthoryear{Baumgartner \bgroup \em et al.\egroup
  }{2001}]{bgf:Lixto}
Robert Baumgartner, Georg Gottlob, and Sergio Flesca.
\newblock Visual information extraction with {Lixto}.
\newblock In {\em Proceedings of the 27th International Conference on Very
  Large Databases}, pages 119--128, Rome, Italy, September 2001. Morgan
  Kaufmann.

\bibitem[\protect\citeauthoryear{Brachman and
  Schmolze}{1985}]{brachman-schmolze:kl-one}
Ronald~J. Brachman and James~G. Schmolze.
\newblock An overview of the {KL-ONE} knowledge representation system.
\newblock {\em Cognitive Science}, 9(2):171--216, April--June 1985.

\bibitem[\protect\citeauthoryear{Gottlob \bgroup \em et al.\egroup
  }{2002}]{gls:hypertrees}
Georg Gottlob, Nicola Leone, and Francesco Scarcello.
\newblock Hypertree decompositions and tractable queries.
\newblock {\em Journal of Computer and System Sciences}, 64(3):579--627, May
  2002.

\bibitem[\protect\citeauthoryear{Gottlob}{1992}]{gottlob:nonmon}
Georg Gottlob.
\newblock Complexity results for nonmonotonic logics.
\newblock {\em Journal of Logic and Computation}, 2(3):397--425, June 1992.

\bibitem[\protect\citeauthoryear{Levesque}{1984a}]{levesque:functional-foundations}
Hector~J. Levesque.
\newblock Foundations of a functional approach to knowledge representation.
\newblock {\em Artificial Intelligence}, 23(2):155--212, July 1984.

\bibitem[\protect\citeauthoryear{Levesque}{1984b}]{levesque:belief}
Hector~J. Levesque.
\newblock A logic of implicit and explicit belief.
\newblock In {\em Proceedings of the Fourth National Conference on Artificial
  Intelligence}, pages 198--202, Austin, Texas, August 1984. American
  Association for Artificial Intelligence.

\bibitem[\protect\citeauthoryear{Nebel}{2000}]{nebel:jair-2000}
Bernhard Nebel.
\newblock On the compilability and expressive power of propositional planning
  formalisms.
\newblock {\em Journal of Artificial Intelligence Research}, 12:271--315, 2000.

\end{thebibliography}


\begin{thebibliography}{}

\bibitem[\protect\citeauthoryear{Ali \bgroup \em et al.\egroup }{2023}]{ali2023deep}
Zafar Ali, Yi~Huang, Irfan Ullah, Junlan Feng, Chao Deng, Nimbeshaho Thierry, Asad Khan, Asim~Ullah Jan, Xiaoli Shen, Wu~Rui, et~al.
\newblock {Deep Learning for Medication Recommendation: A Systematic Survey}.
\newblock {\em Data Intelligence}, 5(2):303--354, 2023.

\bibitem[\protect\citeauthoryear{Almirall \bgroup \em et al.\egroup }{2012}]{pilot}
Daniel Almirall, Scott~N Compton, Meredith Gunlicks-Stoessel, Naihua Duan, and Susan~A Murphy.
\newblock Designing a pilot sequential multiple assignment randomized trial for developing an adaptive treatment strategy.
\newblock {\em Statistics in Medicine}, 31(17):1887--1902, 2012.

\bibitem[\protect\citeauthoryear{Ba \bgroup \em et al.\egroup }{2016}]{layernorm}
Jimmy~Lei Ba, Jamie~Ryan Kiros, and Geoffrey~E Hinton.
\newblock Layer normalization.
\newblock {\em arXiv preprint arXiv:1607.06450}, 2016.

\bibitem[\protect\citeauthoryear{Bajor and Lasko}{2022}]{pmdc-rnn}
Jacek~M Bajor and Thomas~A Lasko.
\newblock Predicting medications from diagnostic codes with recurrent neural networks.
\newblock In {\em Proceedings of the International Conference on Learning Representations (ICLR)}, 2022.

\bibitem[\protect\citeauthoryear{Bhoi \bgroup \em et al.\egroup }{2021}]{premier}
Suman Bhoi, Mong~Li Lee, Wynne Hsu, Hao Sen~Andrew Fang, and Ngiap~Chuan Tan.
\newblock Personalizing medication recommendation with a graph-based approach.
\newblock {\em ACM Transactions on Information Systems (TOIS)}, 40(3):1--23, 2021.

\bibitem[\protect\citeauthoryear{Bhoi \bgroup \em et al.\egroup }{2023}]{refine}
Suman Bhoi, Mong~Li Lee, Wynne Hsu, and Ngiap~Chuan Tan.
\newblock Refine: A fine-grained medication recommendation system using deep learning and personalized drug interaction modeling.
\newblock 36:24013--24024, 2023.

\bibitem[\protect\citeauthoryear{Chen \bgroup \em et al.\egroup }{2016}]{advisory}
Zhuo Chen, Kyle Marple, Elmer Salazar, Gopal Gupta, and Lakshman Tamil.
\newblock A physician advisory system for chronic heart failure management based on knowledge patterns.
\newblock {\em Theory and Practice of Logic Programming}, 16(5-6):604--618, 2016.

\bibitem[\protect\citeauthoryear{Chen \bgroup \em et al.\egroup }{2023}]{carmen}
Qianyu Chen, Xin Li, Kunnan Geng, and Mingzhong Wang.
\newblock Context-aware safe medication recommendations with molecular graph and ddi graph embedding.
\newblock In {\em Proceedings of the AAAI Conference on Artificial Intelligence (AAAI)}, pages 7053--7060, 2023.

\bibitem[\protect\citeauthoryear{Choi \bgroup \em et al.\egroup }{2016a}]{doctorai}
Edward Choi, Mohammad~Taha Bahadori, Andy Schuetz, Walter~F Stewart, and Jimeng Sun.
\newblock Doctor ai: Predicting clinical events via recurrent neural networks.
\newblock In {\em Machine Learning for Healthcare Conference}, pages 301--318, 2016.

\bibitem[\protect\citeauthoryear{Choi \bgroup \em et al.\egroup }{2016b}]{retain}
Edward Choi, Mohammad~Taha Bahadori, Jimeng Sun, Joshua Kulas, Andy Schuetz, and Walter Stewart.
\newblock Retain: An interpretable predictive model for healthcare using reverse time attention mechanism.
\newblock In {\em Advances in Neural Information Processing Systems (NeurIPS)}, volume~29, 2016.

\bibitem[\protect\citeauthoryear{Dosovitskiy and Djolonga}{2019}]{loss_sum}
Alexey Dosovitskiy and Josip Djolonga.
\newblock You only train once: Loss-conditional training of deep networks.
\newblock In {\em International Conference on Learning Representations (ICLR)}, 2019.

\bibitem[\protect\citeauthoryear{Gong \bgroup \em et al.\egroup }{2021}]{smr}
Fan Gong, Meng Wang, Haofen Wang, Sen Wang, and Mengyue Liu.
\newblock Smr: medical knowledge graph embedding for safe medicine recommendation.
\newblock {\em Big Data Research}, 23:100174, 2021.

\bibitem[\protect\citeauthoryear{Gu and Dao}{2023}]{mamba}
Albert Gu and Tri Dao.
\newblock Mamba: Linear-time sequence modeling with selective state spaces.
\newblock {\em arXiv preprint arXiv:2312.00752}, 2023.

\bibitem[\protect\citeauthoryear{He \bgroup \em et al.\egroup }{2020}]{amanet}
Yong He, Cheng Wang, Nan Li, and Zhenyu Zeng.
\newblock Attention and memory-augmented networks for dual-view sequential learning.
\newblock In {\em Proceedings of the 26th ACM SIGKDD International Conference on Knowledge Discovery and Data Mining (KDD)}, pages 125--134, 2020.

\bibitem[\protect\citeauthoryear{Johnson \bgroup \em et al.\egroup }{2016}]{mimic3}
Alistair~EW Johnson, Tom~J Pollard, Lu~Shen, Li-wei~H Lehman, Mengling Feng, Mohammad Ghassemi, Benjamin Moody, Peter Szolovits, Leo Anthony~Celi, and Roger~G Mark.
\newblock Mimic-iii, a freely accessible critical care database.
\newblock {\em Scientific Data}, 3(1):1--9, 2016.

\bibitem[\protect\citeauthoryear{Johnson \bgroup \em et al.\egroup }{2018}]{mimic4}
Alistair~EW Johnson, David~J Stone, Leo~A Celi, and Tom~J Pollard.
\newblock The mimic code repository: enabling reproducibility in critical care research.
\newblock {\em Journal of the American Medical Informatics Association}, 25(1):32--39, 2018.

\bibitem[\protect\citeauthoryear{Lakkaraju and Rudin}{2017}]{markov}
Himabindu Lakkaraju and Cynthia Rudin.
\newblock Learning cost-effective and interpretable treatment regimes.
\newblock In {\em Proceedings of the Artificial Intelligence and Statistics (AISTATS)}, pages 166--175. PMLR, 2017.

\bibitem[\protect\citeauthoryear{Le \bgroup \em et al.\egroup }{2018}]{dmnc}
Hung Le, Truyen Tran, and Svetha Venkatesh.
\newblock Dual memory neural computer for asynchronous two-view sequential learning.
\newblock In {\em Proceedings of the 24th ACM SIGKDD Conference on Knowledge Discovery and Data Mining (KDD)}, pages 1637--1645, 2018.

\bibitem[\protect\citeauthoryear{Liu \bgroup \em et al.\egroup }{2024}]{leader_llm}
Qidong Liu, Xian Wu, Xiangyu Zhao, Yuanshao Zhu, Zijian Zhang, Feng Tian, and Yefeng Zheng.
\newblock Large language model distilling medication recommendation model.
\newblock {\em arXiv preprint arXiv:2402.02803}, 2024.

\bibitem[\protect\citeauthoryear{Shang \bgroup \em et al.\egroup }{2019a}]{gbert}
Junyuan Shang, Tengfei Ma, Cao Xiao, and Jimeng Sun.
\newblock Pre-training of graph augmented transformers for medication recommendation.
\newblock pages 5953--5959, 7 2019.

\bibitem[\protect\citeauthoryear{Shang \bgroup \em et al.\egroup }{2019b}]{gamenet}
Junyuan Shang, Cao Xiao, Tengfei Ma, Hongyan Li, and Jimeng Sun.
\newblock Gamenet: Graph augmented memory networks for recommending medication combination.
\newblock In {\em Proceedings of the AAAI Conference on Artificial Intelligence (AAAI)}, volume~33, pages 1126--1133, 2019.

\bibitem[\protect\citeauthoryear{Solt and Tikk}{2009}]{rulebased}
Ill{\'e}s Solt and Domonkos Tikk.
\newblock Yet another rule-based approach for extracting medication information from discharge summaries.
\newblock In {\em Proceedings of the Third I2B2 Workshop on Challenges in Natural Language Processing for Clinical Data}, 2009.

\bibitem[\protect\citeauthoryear{Tan \bgroup \em et al.\egroup }{2022}]{4sdrug}
Yanchao Tan, Chengjun Kong, Leisheng Yu, Pan Li, Chaochao Chen, Xiaolin Zheng, Vicki~S Hertzberg, and Carl Yang.
\newblock 4sdrug: Symptom-based set-to-set small and safe drug recommendation.
\newblock In {\em Proceedings of the 28th ACM SIGKDD Conference on Knowledge Discovery and Data Mining (KDD)}, pages 3970--3980, 2022.

\bibitem[\protect\citeauthoryear{Wu \bgroup \em et al.\egroup }{2022}]{cognet}
Rui Wu, Zhaopeng Qiu, Jiacheng Jiang, Guilin Qi, and Xian Wu.
\newblock Conditional generation net for medication recommendation.
\newblock In {\em Proceedings of the ACM Web Conference (WWW)}, pages 935--945, 2022.

\bibitem[\protect\citeauthoryear{Xiao \bgroup \em et al.\egroup }{2018}]{ehr_review}
Cao Xiao, Edward Choi, and Jimeng Sun.
\newblock Opportunities and challenges in developing deep learning models using electronic health records data: a systematic review.
\newblock {\em Journal of the American Medical Informatics Association}, 25(10):1419--1428, 2018.

\bibitem[\protect\citeauthoryear{Yang \bgroup \em et al.\egroup }{2021a}]{micron}
Chaoqi Yang, Cao Xiao, Lucas Glass, and Jimeng Sun.
\newblock Change matters: Medication change prediction with recurrent residual networks.
\newblock In {\em Proceedings of the 30th International Joint Conference on Artificial Intelligence (IJCAI)}, pages 3728--3734, 2021.

\bibitem[\protect\citeauthoryear{Yang \bgroup \em et al.\egroup }{2021b}]{safedrug}
Chaoqi Yang, Cao Xiao, Fenglong Ma, Lucas Glass, and Jimeng Sun.
\newblock Safedrug: Dual molecular graph encoders for recommending effective and safe drug combinations.
\newblock pages 3735--3741, 8 2021.

\bibitem[\protect\citeauthoryear{Yang \bgroup \em et al.\egroup }{2023}]{molerec}
Nianzu Yang, Kaipeng Zeng, Qitian Wu, and Junchi Yan.
\newblock Molerec: Combinatorial drug recommendation with substructure-aware molecular representation learning.
\newblock In {\em Proceedings of the ACM Web Conference (WWW)}, pages 4075--4085, 2023.

\bibitem[\protect\citeauthoryear{Zhang \bgroup \em et al.\egroup }{2017}]{leap}
Yutao Zhang, Ruoqi Chen, Jie Tang, Walter~F Stewart, and Jimeng Sun.
\newblock Leap: learning to prescribe effective and safe treatment combinations for multimorbidity.
\newblock In {\em Proceedings of the 23rd ACM SIGKDD Conference on Knowledge Discovery and Data Mining (KDD)}, pages 1315--1324, 2017.

\bibitem[\protect\citeauthoryear{Zhang \bgroup \em et al.\egroup }{2023}]{kgmt}
Yingying Zhang, Xian Wu, Quan Fang, Shengsheng Qian, and Changsheng Xu.
\newblock Knowledge-enhanced attributed multi-task learning for medicine recommendation.
\newblock {\em ACM Transactions on Information Systems (TOIS)}, 41(1):1--24, 2023.

\end{thebibliography}

\end{document}